\documentclass[a4paper,fleqn]{cas-dc}

\usepackage[authoryear,longnamesfirst]{natbib}

\def\tsc#1{\csdef{#1}{\textsc{\lowercase{#1}}\xspace}}
\tsc{WGM}
\tsc{QE}

\begin{document}
\let\WriteBookmarks\relax
\def\floatpagepagefraction{1}
\def\textpagefraction{.001}

\shorttitle{Face inpainting with Identity Preserving Latent Diffusion Models}    

\shortauthors{J. Santos et al.}  

\title []{Face Inpainting with Identity Preserving Latent Diffusion Models}  

\tnotemark[1] 

\tnotetext[1]{} 

%

\author[1]{Jo\~ao Santos}

\affiliation[1]{organization={Institute for Systems and Robotics, Instituto Superior Técnico},
            city={Lisboa},
            country={Portugal}}

\author[1]{Carlos Santiago}

\cormark[1]


\ead{carlos.santiago@tecnico.ulisboa.pt}



\author[1]{Manuel Marques}

\cortext[1]{Corresponding author}



\begin{abstract}
Face inpainting techniques recover missing or occluded facial regions in a visually realistic manner, but preserving the identity in the final output remains a fundamental challenge. Identity consistency is crucial for downstream applications such as face recognition, digital forensics, and human–computer interaction, where even subtle identity distortions can significantly degrade performance or trust. Although diffusion-based generative models have recently achieved remarkable progress in image inpainting, they often struggle to faithfully retain individual-specific facial characteristics. On the other hand, existing identity-aware methods typically rely on costly fine-tuning, auxiliary supervision, or exhibit limited robustness to diverse occlusions, poses, and facial variations.
To address these limitations, we propose ID-ControlNet, an identity-preserving face inpainting framework built upon latent diffusion models. 
Based on ControlNet architecture, our approach conditions the diffusion process on facial identity embeddings extracted from a pretrained face recognition network. 
This design enables reconstruction of occluded facial regions while maintaining global facial coherence and identity fidelity. Furthermore, we introduce an identity consistency and triplet loss training strategy that explicitly enforces alignment between the generated face and the target identity representation. Extensive experiments on CelebA-HQ, FFHQ, and on a new E-Mask dataset demonstrate that ID-ControlNet significantly improves identity preservation over standard diffusion-based inpainting methods, achieving performance comparable to state-of-the-art identity-aware approaches.
\end{abstract}







\maketitle

\section{Introduction}

Face inpainting aims to reconstruct missing or occluded facial regions while producing visually realistic and semantically coherent results. This task plays a critical role in numerous real-world applications, including photo restoration, digital content editing, and virtual or augmented reality. In high-stakes domains such as law enforcement and surveillance analysis, identity-consistent inpainting is particularly important, as reconstructed faces must preserve recognizable features that are essential for accurate identification. Consequently, an effective face inpainting system must go beyond visual plausibility and ensure faithful preservation of an individual’s identity.

Diffusion Models (DM) and their variants \citep{ho2020denoisingdiffusionprobabilisticmodels} have demonstrated strong capability in the generation of real images. Despite these advances, keeping the identity in face inpainting remains a persistent challenge. Standard diffusion models lack explicit mechanisms to encode subject-specific identity information, often causing reconstructed regions to drift toward an average facial appearance. As a result, visually convincing outputs may still fail to maintain the unique identity of the person depicted.

To address this issue, recent identity-preserving inpainting methods have explored both GAN-based and diffusion-based solutions. GAN-based approaches \citep{goodfellow2014generativeadversarialnetworks}, such as MyStyle \citep{nitzan2022mystylepersonalizedgenerativeprior}, achieve strong identity reconstruction by leveraging subject-specific fine-tuning. More recently, diffusion-based personalization frameworks, including Custom Diffusion \citep{kumari2023multiconceptcustomizationtexttoimagediffusion} and PVA \citep{xu2023personalizedfaceinpaintingdiffusion}, have demonstrated that high-quality and identity-consistent face inpainting can be achieved using only a small number of reference images. These works establish diffusion models as the dominant paradigm for identity-aware face generation. However, they still depend on identity-specific adaptation or fine-tuning during inference, which limits scalability and hinders generalization to unseen identities. Moreover, identity consistency is typically enforced only at the output level, without explicitly guiding the diffusion process to maintain alignment with identity representations throughout denoising.

In contrast, we propose an identity-conditioned inpainting framework that integrates facial identity cues directly into the diffusion process. Specifically, we introduce ID-ControlNet, a novel approach that extends a pretrained Latent Diffusion Model \citep{rombach2022highresolutionimagesynthesislatent} by injecting identity embeddings, 
extracted from a face recognition approach~\citep{Deng_2022} via a ControlNet-inspired conditioning architecture \citep{zhang2023addingconditionalcontroltexttoimage}. This design enables subject-specific identity information to condition the denoising trajectory, allowing faithful reconstruction of occluded facial regions while preserving global identity coherence. Furthermore, we introduce an identity consistency and triplet loss training strategy that explicitly reinforces alignment between generated faces and their corresponding identity embeddings in a self-supervised manner, eliminating the need for additional identity-specific fine-tuning.

In addition to methodological challenges, evaluating identity preservation in face inpainting remains nontrivial. Existing datasets such as CelebA-HQ~\citep{karras2018progressivegrowinggansimproved} and FFHQ~\citep{karras2019stylebasedgeneratorarchitecturegenerative} provide high-quality facial images but typically employ random or large-area occlusions, which emphasize overall realism rather than precise recovery of identity-defining features. Such settings make it difficult to assess whether a model truly reconstructs missing identity cues or simply generates plausible facial content. To address this limitation, we introduce the E-Mask dataset, a controlled benchmark where occlusions are restricted to the eye region, one of the most identity-informative facial areas. Each image is paired with a predefined eye-region mask, enabling consistent and focused evaluation of identity restoration under controlled occlusions.
We evaluate our approach on CelebA-HQ, FFHQ, and the proposed E-Mask dataset, demonstrating that ID-ControlNet achieves superior identity preservation compared to standard diffusion-based inpainting methods, while maintaining competitive visual quality and generalization across identities and occlusion patterns.

\section{Related Work}

\paragraph{\textbf{Face Inpainting}} Early approaches based on encoder and decoder or Generative Adversarial Networks (GANs) frameworks, such as Context Encoder \citep{pathak2016contextencodersfeaturelearning} and Partial Convolutions \citep{liu2018imageinpaintingirregularholes}, introduced contextual reasoning to fill missing regions. Later improvements like EdgeConnect \citep{nazeri2019edgeconnectgenerativeimageinpainting} and Gated Convolution \citep{yu2019freeformimageinpaintinggated} enhanced structural consistency and mask adaptability. While these models achieve visually convincing completions, they often disregard subject-specific features, resulting in reconstructions that deviate from the individual’s identity when identity-critical regions (e.g., the eyes or nose) are occluded.
\paragraph{\textbf{Identity and Reference-Guided Inpainting}} To improve identity consistency, several subsequent works integrated explicit identity priors or reference images. 
ID-Inpainter \citep{zou2025videovirtualtryonconditional} incorporated embeddings from pretrained face recognition networks to guide identity recovery. MyStyle \citep{nitzan2022mystylepersonalizedgenerativeprior} fine-tuned a StyleGAN~\citep{karras2019stylebasedgeneratorarchitecturegenerative} model per individual, learning a personalized latent prior for identity-accurate synthesis, but at high computational cost. Reference-Guided Inpainting \citep{liu2022referenceguidedtexturestructureinference} introduced modular control for texture and structure using adaptive normalization, while PATMAT \citep{motamed2023patmatpersonawaretuning} employed person-specific transformer anchors for large-occlusion restoration. 

These approaches improved fine-grained fidelity yet rely heavily on per-identity adaptation or multiple references. E2F-Net \citep{hassanpour2024e2fneteyestofaceinpaintingstylegan} specifically leveraged the eye region to reconstruct full faces, revealing the discriminative power of periocular cues for identity recovery. However, most of these GAN-based methods struggle to scale to unseen identities without retraining or fine-tuning.
\paragraph{\textbf{Diffusion-Based Personalization and Inpainting}} Diffusion models have recently surpassed GANs in generative fidelity and stability. Latent Diffusion Inpainting \citep{rombach2022highresolutionimagesynthesislatent} and RePaint \citep{lugmayr2022repaintinpaintingusingdenoising} established strong baselines for structural and texture restoration through iterative denoising and personalized diffusion models then extended this paradigm by conditioning generation on specific identity features.

DreamBooth \citep{ruiz2023dreamboothfinetuningtexttoimage} and Custom Diffusion \citep{kumari2023multiconceptcustomizationtexttoimagediffusion} enabled few-shot personalization but still required fine-tuning per subject. Parallel Visual Attention (PVA) \citep{xu2023personalizedfaceinpaintingdiffusion} introduced reference-conditioned attention layers to encode identity directly during the diffusion process, achieving identity-consistent inpainting with minimal adaptation. More recently, IDiff-Face \citep{boutros2023idifffacesyntheticbasedfacerecognition} adapted diffusion modeling for synthetic face generation and recognition, employing identity-conditioned latent diffusion with contextual dropout to balance identity discrimination and intra-class variation. These advances position diffusion-based frameworks as the new state of the art for personalized face generation, yet challenges remain in integrating identity conditioning efficiently without subject-specific optimization.

\section{Background}

This section provides a brief overview of Diffusion Models and ControlNet\citep{zhang2023addingconditionalcontroltexttoimage}, which form the foundational components of our approach.

Diffusion Probabilistic Models (DPMs) \citep{ho2020denoisingdiffusionprobabilisticmodels} are a class of generative models that synthesize data by learning to reverse a gradual noising process. Starting from a clean image, $x_0 \sim p_{\mathrm{data}}(x)$, the forward diffusion process progressively corrupts the data with Gaussian noise over timesteps, $t = 1,\dots,T$, according to
\begin{equation}
q(x_t \mid x_{t-1}) =
\mathcal{N}\left(\sqrt{1-\beta_t},x_{t-1},,\beta_t \mathbf{I}\right),
\end{equation}
where $\beta_t \in (0,1)$ controls the noise schedule. This process admits a closed-form expression with respect to the original image,
\begin{equation}
q(x_t \mid x_0)
= \mathcal{N}\!\left(\sqrt{\bar{\alpha}_t},x_0,,
(1-\bar{\alpha}_t)\mathbf{I}\right),
\quad
\bar{\alpha}_t=\prod_{s=1}^{t}(1-\beta_s).
\end{equation}

The generative task is then formulated as learning the reverse process, which iteratively denoises a noisy sample back to a clean image. This reverse transition is parameterized by a neural network:
\begin{equation}
p_\theta(x_{t-1}\mid x_t)
= \mathcal{N}\!\left(
\mu_\theta(x_t,t),
\Sigma_\theta(x_t,t)
\right).
\end{equation}

The model is trained to predict the added noise ($\epsilon$) at each timestep using the noise-prediction objective
\begin{equation}
\mathcal{L}_{\text{DM}}
= \mathbb{E}_{x_0,\epsilon,t}
\left[
|\epsilon -
\epsilon_\theta(x_t,t)|_2^2
\right],
\end{equation}
where
$x_t = \sqrt{\bar{\alpha}_t}x_0 + \sqrt{1-\bar{\alpha}_t},\epsilon,
\quad \epsilon \sim \mathcal{N}(0,\mathbf{I})$ and $\epsilon_\theta$ is the model based on the U-Net architecture.
At inference time, generation begins from pure noise $x_T \sim \mathcal{N}(0,\mathbf{I}$ and progressively applies the learned denoising steps to obtain a sample $x_0$.

Although diffusion models achieve excellent visual fidelity and sample diversity, operating directly in pixel space is computationally expensive, particularly for high-resolution images.
Latent Diffusion Models (LDMs) \citep{rombach2022highresolutionimagesynthesislatent} alleviate this computational burden by performing diffusion in a compact latent space learned by a variational autoencoder (VAE). An encoder $\mathcal{E}$ maps an image $x_0$ to a latent representation $z_0 = \mathcal{E}(x_0)$, and a decoder $\mathcal{D}$ reconstructs the image from the latent code, $\hat{x}_0 = \mathcal{D}(z_0)$. Diffusion is then applied to the latent variables
\begin{equation}
q(z_t \mid z_0)
= \mathcal{N}\!\left(
\sqrt{\bar{\alpha}_t}z_0,
(1-\bar{\alpha}_t)\mathbf{I}
\right).
\end{equation}

In a more useful way, the denoising U-Net $\epsilon_\theta(z_t, t, c)$ predicts the noise in latent space and can be conditioned on auxiliary information $c$, such as text embeddings or image context. For inpainting, the conditioning typically includes a masked image, $x_{\mathrm{mask}}$, and a binary mask, $m$,
\begin{equation}
z_{\mathrm{cond}} = \mathcal{E}(x_{\mathrm{mask}}), \qquad
\epsilon_\theta(z_t, t, m, z_{\mathrm{cond}}),
\end{equation}
ensuring that known regions remain unchanged while missing areas are synthesized in a manner consistent with the visible context. 

ControlNet \citep{zhang2023addingconditionalcontroltexttoimage} extends pretrained diffusion models by introducing an auxiliary network branch that injects external conditioning signals without disrupting the original generative prior. Let $f_\theta$ denote the frozen backbone diffusion U-Net and $f_\phi^{\mathrm{ctrl}}$ the trainable control network. At each layer $\ell$, control features are integrated into the backbone through zero-initialized convolutions
\begin{equation}
h_\ell' = h_\ell + g_\ell\!\left(f_\phi^{\mathrm{ctrl}}(c_\ell)\right),
\end{equation}
where $h_\ell$ represents the backbone feature map, $c_\ell$ is the control input at the corresponding resolution, and $g_\ell$ is a learnable projection. By training only the control branch, ControlNet enables precise conditioning with minimal data and avoids catastrophic forgetting of the pretrained diffusion model.



In this work, we build upon these principles by extending ControlNet to semantic identity conditioning. Rather than relying on geometric or structural cues, our method injects identity embeddings extracted from a pretrained face recognition network, enabling identity-aware guidance throughout the diffusion process.

\begin{figure*}[!t]
\centering
\includegraphics[width=.75\textwidth]{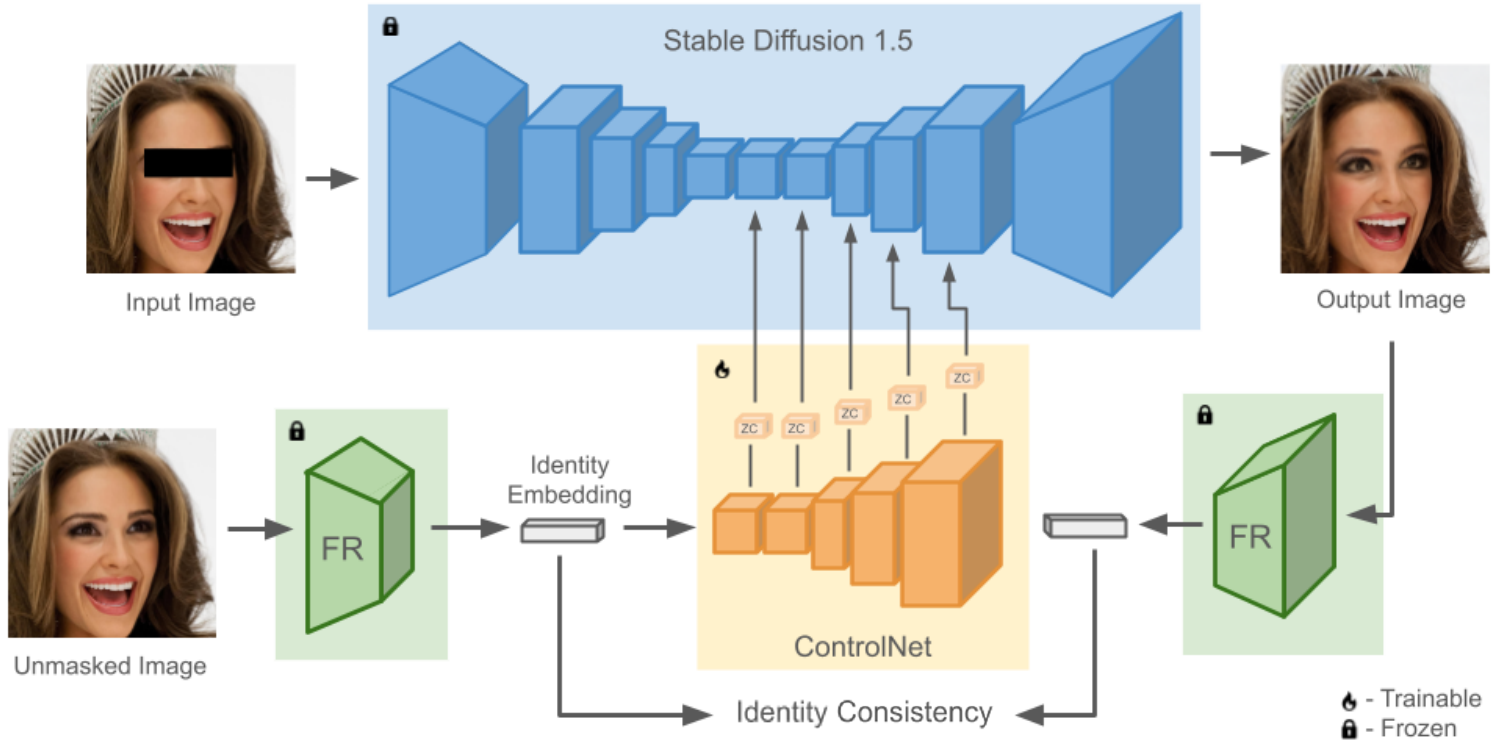}
\caption{Overview of the proposed ID-ControlNet architecture. 
    A frozen face-recognition encoder (FR) extracts an identity embedding from the unmasked reference image, which is projected into spatial control maps by a trainable ControlNet branch. These maps modulate a frozen Stable Diffusion 1.5 inpainting backbone during denoising, guiding the reconstruction toward the target identity. Identity consistency supervision on embedding-level losses enforce alignment between the generated and reference identities. Locked icons indicate frozen components, while flame icons denote trainable modules.}
\label{fig:framework}
\end{figure*}

\section{ID-ControlNet: Identity-Conditioned ControlNet}
Given a masked face image and an associated identity representation extracted from a pretrained face recognition model, the goal of identity-preserving face inpainting is to reconstruct missing regions such that the resulting image remains faithful to the subject’s identity while maintaining high visual realism. 

To address these challenges and as illustrated in Fig. \ref{fig:framework}, we propose ID-ControlNet, an identity-conditioned extension of ControlNet that injects subject-specific identity information directly into a latent diffusion U-Net. During training, only the control branch is optimized, while both the diffusion backbone and the face recognition network (ArcFace, in our implementation) remain frozen. In contrast to the original ControlNet, which enforces spatial conditioning signals (e.g., edges or depth maps) by replicating the U-Net encoder structure, our method introduces identity guidance in the latent space via identity embeddings and integrates it into the decoder pathway.

\subsection{Identity Embeddings}

In ControlNet~\citep{zhang2023addingconditionalcontroltexttoimage}, the control branch replicates the downsampling path of the U-Net, processing a $64\times64$ input and progressively encoding it into an $8\times8$ latent representation, with intermediate feature maps injected into the corresponding backbone layers. In contrast, our approach operates directly in the latent space, where identity embeddings are defined. Specifically, the decoder block upsamples the identity embeddings from $8\times8$ back to $64\times64$, ensuring that the control branch remains aligned with the same connection points as in the original ControlNet architecture.

At each resolution level, zero-convolution layers inject identity-guided features into the U-Net model of Stable Diffusion~\citep{rombach2022highresolutionimagesynthesislatent}, preserving the residual conditioning mechanism of ControlNet while adapting it to compact identity embeddings rather than spatial control signals.

During generation, the Stable Diffusion U-Net progressively denoises the latent representation under the joint guidance of textual prompts and identity embeddings. In the final stage, the generated image is passed through the same frozen face recognition (FR) encoder to extract its identity embedding. This enables the use of an identity consistency loss in the embedding space, whereby the similarity between the conditioning embedding and the embedding of the generated image is maximized, promoting identity preservation.

\subsection{Training}

Training is carried out on a CelebA-HQ-IDI subset (see details in Section~\ref{sec:experiments}) and uses a composite objective that combines the LDM denoising loss with embedding-level supervision. The denoising term follows the latent diffusion formulation and trains the denoiser to predict the noise component from a noisy latent,
\begin{equation}
\mathcal{L}_{\mathrm{denoise}} \;=\; \mathbb{E}_{x_0,\epsilon,t}\!\left[\big\|\epsilon - \hat\epsilon_\theta\left(z_t,t,m,\,C(e_{\mathrm{cond}})\right)\big\|_2^2\right],
\end{equation}
where $\hat\epsilon_\theta(\cdot)$ is the denoiser prediction conditioned on the mask $m$, and the control maps $C(e_{\mathrm{cond}})$ are obtained from the identity embedding $e_{\mathrm{cond}}$.

Identity fidelity is enforced in the identity embedding space. For each sample $i$ in a batch of $N$, after decoding a generated latent to a generated image $\hat{x}_0^{(i)}$, the frozen recognition encoder computes an embedding, $e_\mathrm{gen}^{(i)}$, that is compared to the conditioning embedding (the external embedding used as input) through an identity consistency loss,
\begin{equation}
\mathcal{L}_{\mathrm{id}} \;=\; \frac{1}{N}\sum_{i=1}^N d(e_{\mathrm{gen}}^{(i)},e_{\mathrm{cond}}^{(i)})\;,
\end{equation}
where $d(u,v)= 1 - u^Tv$ is the cosine distance.
To increase inter-subject separability, encouraging the model to favor the target identity over others, a batch-level triplet loss is added, 
\begin{equation}
\mathcal{L}_{\mathrm{trip}} \;=\; \frac{1}{B}\sum_{i=1}^{B}\max\!\big(0,\; d(e^{(i)}_{\mathrm{gen}},e^{(i)}_{\mathrm{pos}}) - d(e^{(i)}_{\mathrm{gen}},e^{(i)}_{\mathrm{neg}}) + m\big),
\end{equation}
where $e^{(i)}_{\mathrm{pos}}=e_{\mathrm{cond}}^{(i)}$ is the conditioning embedding for sample $i$, $e^{(i)}_{\mathrm{neg}}$ the embedding of a random negative sample from the same batch, and $m$ the margin.


The complete training loss is a weighted sum of all terms
\begin{equation}
\mathcal{L}_{\mathrm{total}} \;=\; \mathcal{L}_{\mathrm{denoise}} + \lambda_{\mathrm{id}}\mathcal{L}_{\mathrm{id}} + \lambda_{\mathrm{trip}}\mathcal{L}_{\mathrm{trip}},
\end{equation}
where coefficients $\lambda_\mathrm{id}$, $\lambda_\mathrm{trip}$ are hyperparameters that are tuned to balance identity enforcement against visual realism.


Since backpropagating the embedding losses through a full $T$-step denoising chain is computationally expensive, we obtain a tractable gradient for embedding supervision adopting a single-step denoising approximation similar to ControlNet++~\citep{li2024controlnetimprovingconditionalcontrols}: the target latent is perturbed to a small timestep $t_s$, one denoising step is applied, the partially denoised latent is decoded to $\hat{x}_0$, and the frozen recognition encoder computes $e_\mathrm{gen}$ used by the embedding losses. This single-step proxy provides a training signal for identity alignment while keeping memory and compute affordable on a single GPU setup.

\section{E-Mask Dataset}

Although semantic masks are available in well-known datasets such as CelebA-HQ-IDI, their effectiveness is limited by the presence of random and inconsistent occlusions. To address this limitation, we introduce a new dataset specifically designed to focus on facial regions most relevant to identity—namely the eyes, nose, and mouth.


The dataset is built from the original CelebA-HQ images by extracting 3D facial keypoints using MediaPipe FaceMesh \citep{kartynnik2019realtimefacialsurfacegeometry}. For each identity-relevant region, the corresponding keypoints are used to define rectangular bounding boxes. Controlled padding is applied to the eye region to ensure robust coverage under pose variations and prevent partial occlusion due to tight cropping. In contrast, the nose and mouth regions require no additional padding, as their bounding boxes provide stable coverage. To ensure consistency, the bounding boxes are defined to be non-overlapping, allowing each mask to isolate a distinct facial region. 

\begin{figure}
    \centering
    \includegraphics[width=.8\linewidth]{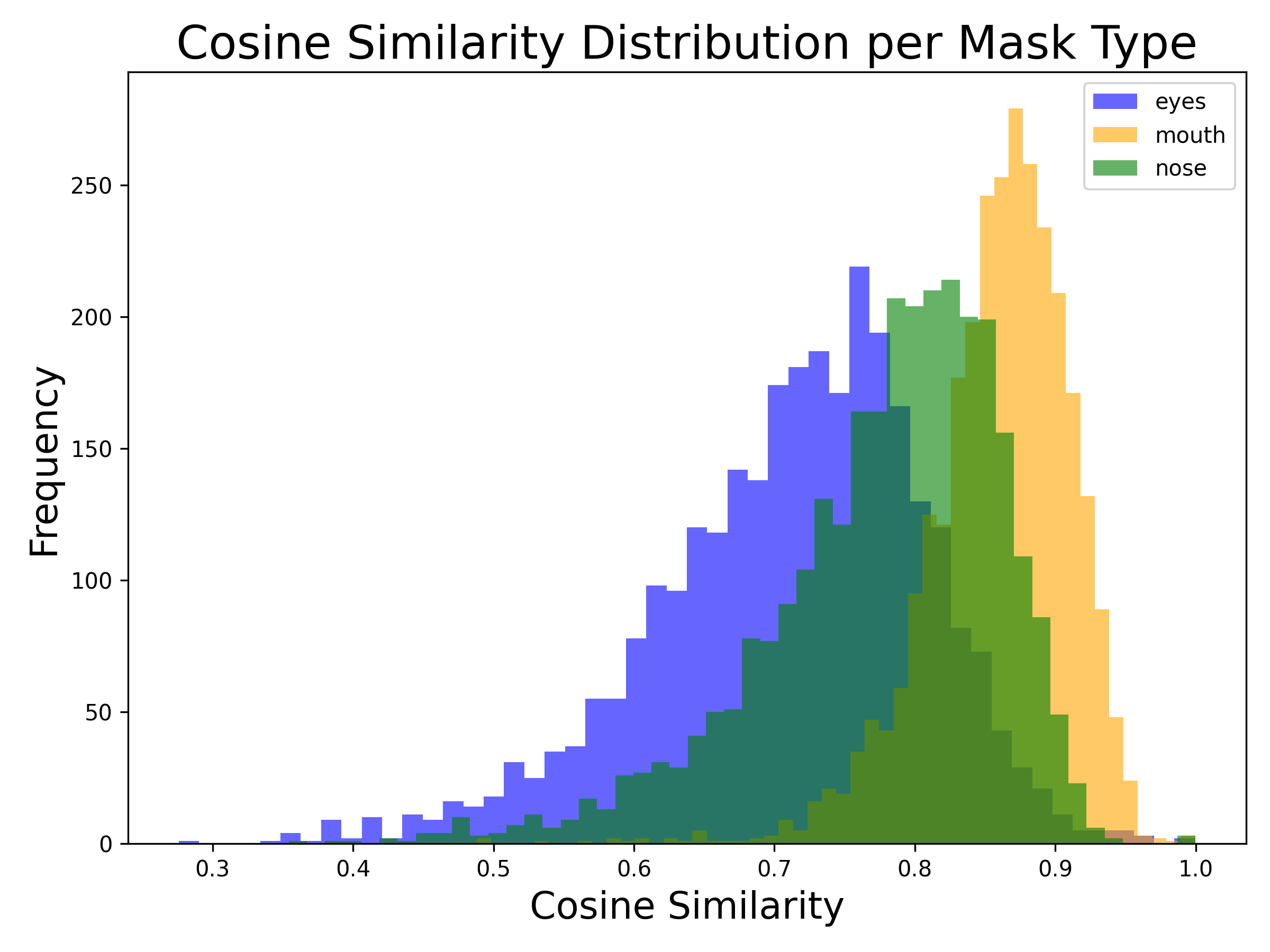}
    \caption{Distribution of cosine similarity scores per mask type.}
    \label{fig:similarity_distribution}
\end{figure}

Figure \ref{fig:similarity_distribution} illustrates the effect of different mask types on identity concealment. The distributions of cosine similarity show that eye masks achieve the strongest suppression of identity, followed by the nose, while mouth masks consistently have the weakest effect. This pattern suggests that the regions around the eyes and nose contain the most identity-relevant features, whereas the mouth contributes comparatively less to facial recognition.

\begin{figure}
    \centering
    \includegraphics[width=.8\linewidth]{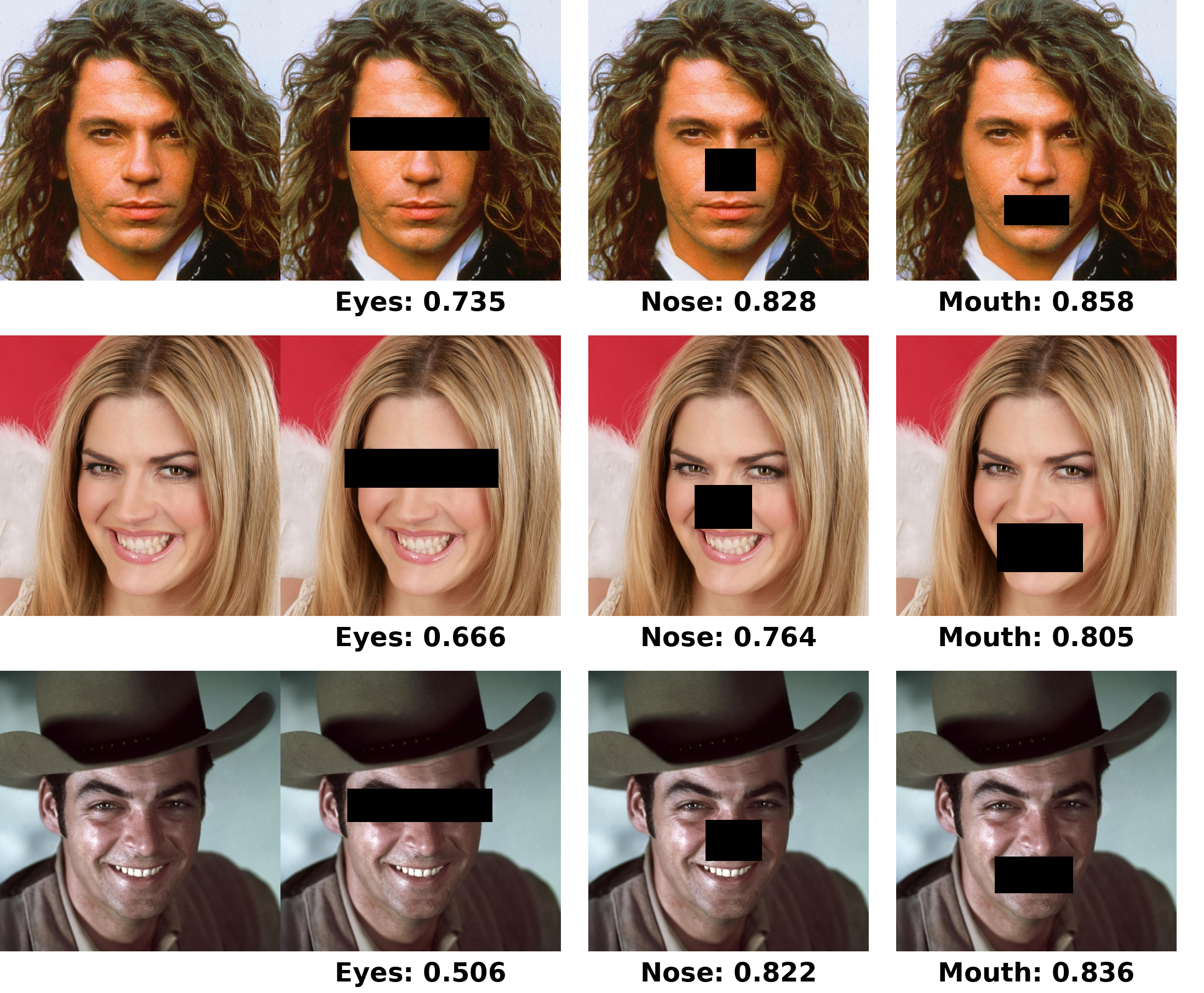}
    \caption{Examples of masked images and their cosine similarity scores with respect to the unmasked originals.}
    \label{fig:mask_examples}
\end{figure}

Complementing these quantitative results, Figure \ref{fig:mask_examples} shows qualitative examples of masked images with their corresponding cosine similarity scores. These visualizations provide an intuitive sense of how different regions impact recognition, confirming that occlusions in the eyes and nose consistently degrade identity features more strongly than nose occlusions, and both are noticeably more effective than the mouth.

Based on these findings, the impact on identity concealment is the decisive factor and, in this regard, the eyes stand out clearly. Beyond their quantitative advantage, the eyes also represent the most practical choice for real-world scenarios, since they are naturally and frequently occluded by glasses or sunglasses, and occasionally by hats, making them a region where identity suppression is more contextually realistic. For these reasons, the E-Mask dataset was ultimately structured around the eye region, ensuring that it emphasizes the most critical and practically relevant area for identity concealment.

\section{Experiments}
\label{sec:experiments}
The experimental framework was implemented in PyTorch and executed on a workstation equipped with an NVIDIA RTX A6000 GPU, an Intel® Xeon® Platinum 8260 CPU @ 2.40 GHz (96 cores), and 376 GB of RAM. All models were trained with a batch size of 4 using the Adam optimizer \citep{kingma2017adam}, with a learning rate of $\eta = 5 \times 10^{-6}$.


\subsection{Pretrained models}

We build on the publicly available Stable Diffusion v1.5 checkpoint as the frozen generative backbone. Identity embeddings are obtained from ArcFace~\citep{Deng_2022} R100 face recognition network trained on MS1MV3~\citep{Deng_2019_ICCV}. To evaluate our approach against another state-of-the-art method, we use a different pretrained network, the CosFace R100 networK~\citep{wang2018cosfacelargemargincosine}  trained on Glint360K~\citep{an2022killingbirdsstoneefficientrobust}.

\subsection{Datasets} 

Experiments are conducted on CelebA-HQ-IDI benchmark~\citep{xu2023personalizedfaceinpaintingdiffusion} and on our identity-focused E-Mask dataset. CelebA-HQ-IDI provides a version of the original CelebA-HQ dataset with identity-aligned splits and reference metadata used for personalized inpainting evaluations. 
We also used the FFHQ dataset~\citep{karras2019stylebasedgeneratorarchitecturegenerative} as an unseen testbed to assess generalization.

All images are center-cropped to the face region, normalized to the input range expected by the latent diffusion backbone, and resized to $512\times512$ for the diffusion pipeline. Identity embeddings are $l_2$-normalized and computed from the original (unmasked) face crops, at the default input resolution of each recognition model; these embeddings are provided during training and evaluation.

\subsection{Evaluation} 

Performance is evaluated with perceptual and identity metrics. Identity preservation is measured as the cosine similarity between ArcFace R100 embeddings of the generated and ground-truth (unmasked) images. We also report the cosine similarity computed with CosFace R100 \citep{wang2018cosfacelargemargincosine}. Perceptual quality is measured with Fréchet Inception Distance (FID)~\citep{heusel2018ganstrainedtimescaleupdate}, Kernel Inception Distance (KID)~\citep{binkowski2021demystifyingmmdgans}, masked FID (mFID) computed only over the inpainted region, and LPIPS~\citep{zhang2018unreasonableeffectivenessdeepfeatures}.

For each test image, the model receives a masked input and its corresponding identity embedding extracted from the unmasked version. Reconstructions are generated using the same sampling schedule and number of denoising steps across all models. To assess generalization, models trained on CelebA-HQ-IDI are evaluated directly on the unseen FFHQ dataset without additional fine-tuning.

\section{Results} 

In this section, we evaluate our methodology in terms of perceptual quality and identity preservation, using CelebA-HQ and E-Mask datasets in both assessments.

\subsection{Perceptual Quality}

To evaluate the impact of identity conditioning in image quality, we compared ID-ControlNet against the baseline SD1.5 across CelebA-HQ and FFHQ test sets.
Table~\ref{tab:celeba_quality_agg} reports these results, where both models achieve comparable global image quality. 
Similar results were obtained on E-Mask, under eye-specific occlusions, as shown in Table~\ref{tab:em_quality}.

\begin{table}[t]
\centering
\caption{Inpainting perceptual quality on CelebA-HQ.
}
\label{tab:celeba_quality_agg}
\renewcommand{\arraystretch}{1.4}
\setlength{\tabcolsep}{8pt}
\scriptsize
\begin{tabular}{lccc}
\hline
Method & FID $\downarrow$ & mFID $\downarrow$ & LPIPS $\downarrow$ \\
\hline
SD1.5 & 14.25 & 18.32 & 0.55 \\
ID-ControlNet         & \textbf{14.13} & \textbf{18.10} & \textbf{0.54} \\
\hline
\end{tabular}
\end{table}

\begin{table}[t]
\centering
\caption{Inpainting perceptual quality on E-Mask.}
\label{tab:em_quality}
\renewcommand{\arraystretch}{1.3}
\setlength{\tabcolsep}{11pt}
\scriptsize
\begin{tabular}{lccc}
\hline
Method & FID $\downarrow$ & mFID $\downarrow$ & LPIPS $\downarrow$ \\
\hline
SD1.5 & \textbf{13.35} & 19.58 & \textbf{0.55} \\
ID-ControlNet & 13.45 & \textbf{19.50} & \textbf{0.55} \\
\hline
\end{tabular}
\end{table}


Comparing both datasets, E-Mask metrics are lower than those obtained on CelebA-HQ-IDI. However, these differences are not substantial, suggesting that both models behave similarly across different masking conditions. This indicates that identity conditioning does not significantly affect perceptual quality under the evaluated settings.

\subsection{Identity Preservation}

We compare our approach with other inpainting strategies: SD2~\citep{rombach2022highresolutionimagesynthesislatent}, Paint by Example~\citep{yang2023paint}, MyStyle~\citep{nitzan2022mystylepersonalizedgenerativeprior}, Textual Inversion~\citep{gal2023image}, Custom Diffusion~\citep{kumari2023multiconceptcustomizationtexttoimagediffusion}, and PVA~\citep{xu2023personalizedfaceinpaintingdiffusion}, under two settings: on CelebA-HQ-IDI-5 and on our E-Mask.

\paragraph{\textbf{CelebA-HQ-IDI-5}}
In Table~\ref{tab:pva_comparison}, we evaluate ID-ControlNet through the original PVA benchmark, on a shared subset of 320 identities, each with three semantic masks (960 images per method). All models use identical inputs and preprocessing. For each generated image, identity similarity is computed as the cosine similarity between CosFace~R100 embeddings of the inpainted and ground-truth faces.
Importantly, our approach operates in a zero-shot setting, unlike the fine-tuned PVA variant that adapts per identity using five reference images. In “PVA (No-Tune)”, we report the results obtained using PVA without per-subject adaptation, which represents a fairer comparison. In this setting, ID-ControlNet matches PVA’s identity fidelity while maintaining competitive image quality, despite using an older Stable Diffusion backbone (SD1.5 vs SD2).

\begin{table}[t]
\centering
\caption{Comparison with other identity-aware inpainting methods on CelebA-HQ-IDI-5. Table adapted from~\citep{xu2023personalizedfaceinpaintingdiffusion}.}
\label{tab:pva_comparison}
\scriptsize
\renewcommand{\arraystretch}{1.1}
\setlength{\tabcolsep}{2pt}
\begin{tabular}{lcccc}
\hline
Method & FT.\ Time & ID ↑ & FID ↓ & KID ↓$\times 10^{-3}$ \\
\hline
SD2                    & - & 0.359 & 8.24 & 2.717 \\
Paint by Example       & - & 0.430 & 11.2 & 6.089 \\
MyStyle                & $\sim$15 min & 0.696 & 27.7 & 5.029 \\
Textual Inversion      & $\sim$6 h & 0.644 & 13.8 & 8.404 \\
Custom Diffusion       & $\sim$3 h & 0.729 & 13.9 & 5.870 \\
PVA                    & $\sim$1 min & 0.741 & 8.22 & 4.289 \\
\hline
SD1.5           & - & 0.385 & 28.85 & 3.813 \\
PVA (No-Tune) (SD2)   & - & 0.619 & 19.42 & 1.100 \\
ID-ControlNet (SD1.5) & - & 0.611 & 22.59 & 1.947 \\
\hline
\end{tabular}
\end{table}



\paragraph{\textbf{E-Mask}}

On our eye-specific benchmark, the results, in Table~\ref{tab:em_pva_results}, show that our ID-ControlNet achieves a substantially higher identity score, compared to both the baseline and PVA. For this dataset, SD2 appears to produce better reconstruction quality(FID, mFID, LPIPS).

\begin{figure}
    \centering
    \includegraphics[width=.8\linewidth]{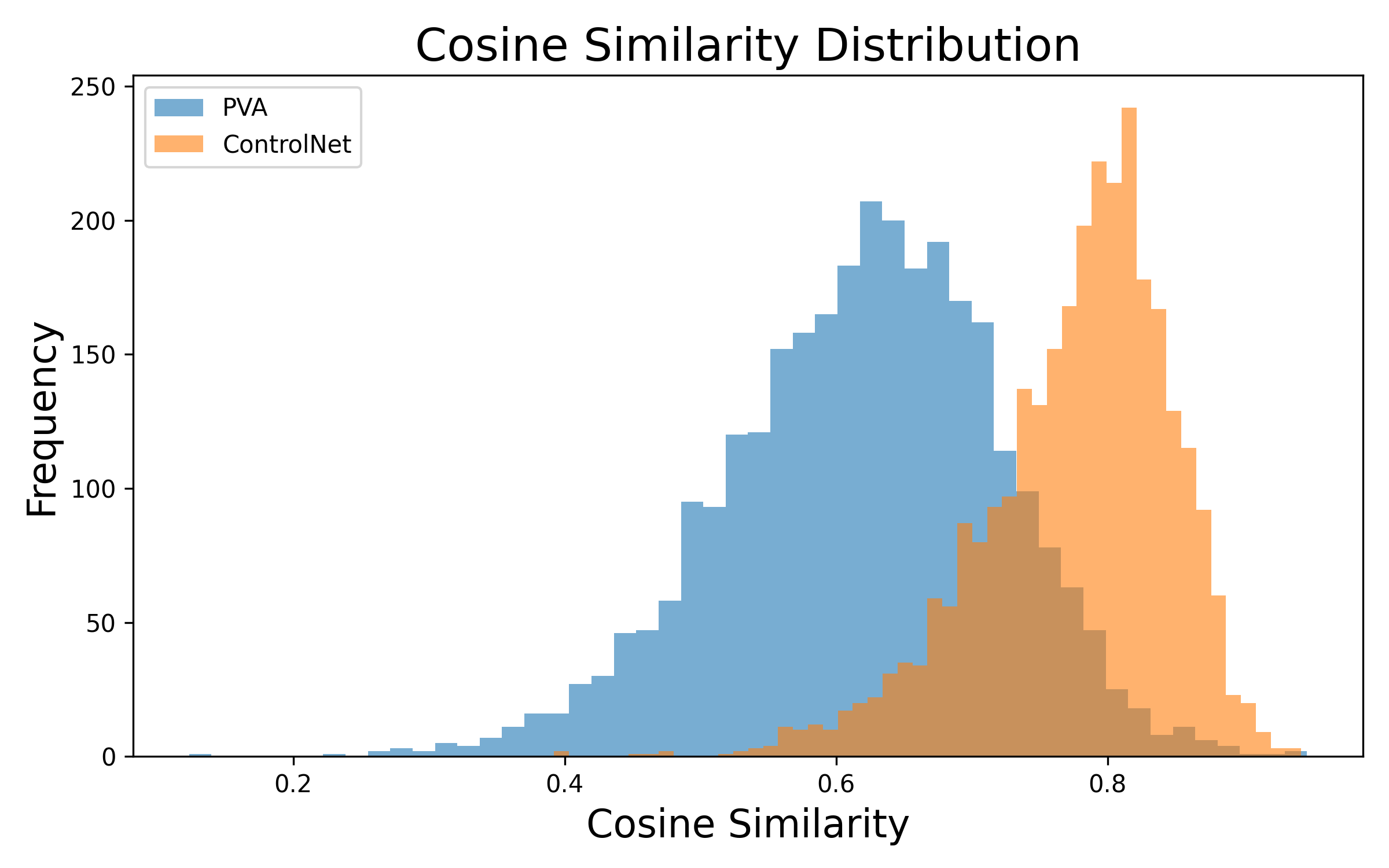}
    \caption{Distribution of identity similarities on E-Mask (CelebA-HQ test set) for PVA (No-Tune) and ID-ControlNet. Higher values indicate stronger identity preservation.}
    \label{fig:em_pva_delta}
\end{figure}

\begin{table}[t]
\centering
\caption{Identity and perceptual quality metrics on E-Mask. Identity similarity is measured via ArcFace cosine similarity.}
\label{tab:em_pva_results}
\scriptsize
\begin{tabular}{lcccc}
\hline
Method  & ID (↑) & FID (↓) & mFID (↓) & LPIPS (↓) \\
\hline
Baseline (SD1.5)        & 0.62 & 13.4 & 19.6 & 0.55 \\
PVA (No-Tune) (SD2)   & 0.63 & \textbf{6.6} & \textbf{3.8} & \textbf{0.04} \\
ID-ControlNet (SD1.5)   & \textbf{0.78} & 13.5 & 19.5 & 0.55 \\
\hline
\end{tabular}
\end{table}

\begin{figure}
    \centering
    \includegraphics[width=1\linewidth]{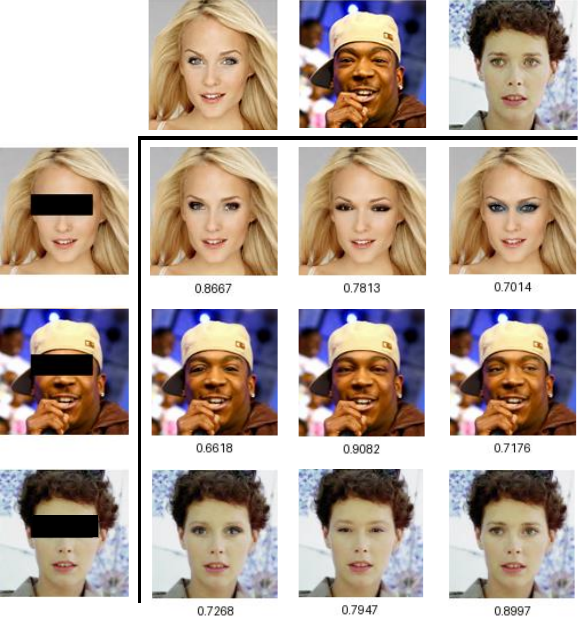}
    \caption{Outcomes of our ID-ControlNet for different combinations of masked images and identity embeddings. Cosine similarity is reported below each result.}
    \label{fig:mask_combinations}
\end{figure}

\subsection{Impact of Identity}

To illustrate the impact of different identity embeddings during inpainting with our approach, we performed inpaint on all combinations of image and identity embedding using three random celebrities from the E-Mask dataset.
Figure~\ref{fig:mask_combinations}, shows that different identity embeddings may lead to substantially different outcomes, but the best identity similarity is obtained for the correct match (along the diagonal).

\begin{figure}
    \centering
    \includegraphics[width=1\linewidth]{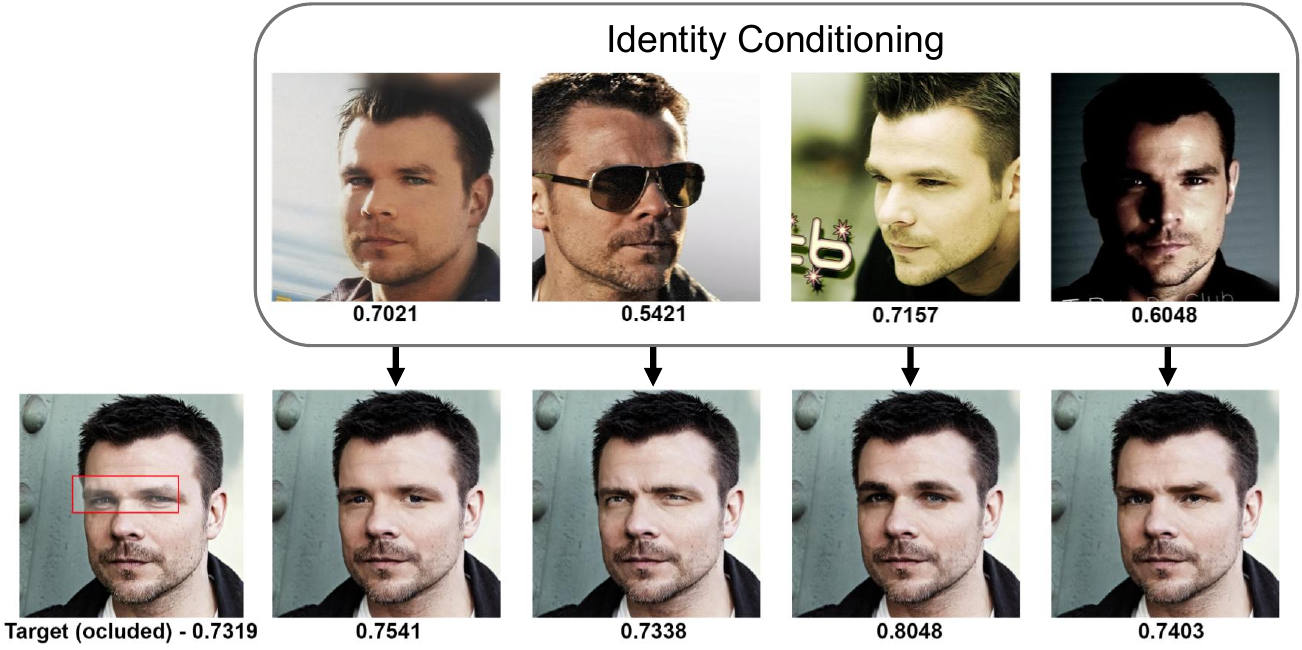}
    \caption{Outcomes of our ID-ControlNet for different identity conditioning from the same person. Cosine similarity is reported below each result.}
    \label{fig:self_combinations}
\end{figure}

Additionally, Figure~\ref{fig:self_combinations} shows the inpainting results when conditioned on identity embeddings obtained from different images of the same person. These results lead to key findings: 1) the quality of the generated image, according to the cosine similarity between identity embeddings, increases with the quality of the conditioning image; 2) all inpainting results improve the cosine similarity between the generated images and the real one (unmasked), even if the conditioning image has low cosine similarity (see, e.g., second identity conditioning).

\section{Conclusion}

We presented ID-ControlNet, a lightweight ControlNet adaptation that injects compact face-identity embeddings into a frozen latent diffusion backbone to produce identity-consistent face inpainting without per-subject fine-tuning. Results show that our approach provides a control over the perceived identity of the reconstructed image, without compromising image quality. Additionally, in zero-shot inpainting, ID-ControlNet outperforms the state of the art in identity preservation. Future work should target backbone upgrades as well as user studies for a more robust assessment of image quality and identity preservation.

\section*{Acknowledgements}
This work is funded by LARSyS funding (DOI: \nolinkurl{10.54499/LA/P/0083/2020}, \nolinkurl{10.54499/UIDP/50009/2020}, and \nolinkurl{10.54499/UIDB/50009/2020}), through Fundação para a Ciência e a Tecnologia. C. Santiago and M. Marques are also supported by the PT Smart Retail project (PRR - \nolinkurl{02/C05-i11/2024.C645440011-00000062}), through IAPMEI - Agência para a Competitividade e Inovação.
\printcredits

\bibliographystyle{cas-model2-names}

\bibliography{biblio}



\end{document}